\documentclass[twoside,twocolumn,10pt]{article}

\usepackage{wscg}           
\RequirePackage{ifpdf}
\ifpdf
 \RequirePackage[pdftex]{graphicx}
 \RequirePackage[pdftex]{color}
\else
 \RequirePackage[dvips,draft]{graphicx}
 \RequirePackage[dvips]{color}
\fi
\usepackage{cite}
\usepackage{amsmath,amssymb,amsfonts}
\usepackage{algorithmic}
\usepackage{textcomp}
\usepackage{xcolor}
\usepackage[binary-units]{siunitx}
\newcommand{\abs}[1]{\left| #1 \right|}
\usepackage{xspace}
\usepackage[tight,footnotesize]{subfigure}
\graphicspath{{pics/}}

\newcommand{\figu}{Figure\xspace}

\newlength{\mylenth}
\usepackage{array}          
\usepackage{tabularx}      
\usepackage{booktabs}      
\usepackage{hyphenat}      

\newif\ifanonymous
\anonymousfalse

\ifanonymous
    \usepackage{lineno}
    \linenumbers
\else
	\usepackage{nopageno}       
\fi

\title{POSTER: Fast and Precise Binary Instance Segmentation of 2D Objects for	Automotive Applications}

\ifanonymous
\else
	\newlength{\affilationW}
	\newlength{\affilationH}
    \author{
    \hspace{-5mm}
        \parbox{\affilationW}{\centering
        Ganganna Ravindra, Darshan\\[1mm]
        CMORE Automotive GmbH\\
        Germany, D-88131 Lindau, Bavaria\\[1mm]
        darshangr4293@\\gmail.com
        }
        \hspace{\affilationH}
        \parbox{\affilationW}{\centering
        Dinges, Laslo\\[1mm]
        Otto-von-Guericke-University\\
        Germany, D-39016 Magdeburg, Saxony-Anhalt \\[1mm]
        Laslo.Dinges@ovgu.de
        }
        \hspace{\affilationH}
        \parbox{\affilationW}{\centering
        Ayoub, Al-Hamadi\\[1mm]
        Otto-von-Guericke-University\\
        Germany, D-39016 Magdeburg, Saxony-Anhalt \\[1mm]
        Ayoub.Al-Hamadi@ovgu.de
        }
        \hspace{\affilationH}
        \parbox{\affilationW}{\centering
        Baranau, Vasili\\[1mm]
        CMORE Automotive GmbH\\
        Germany, D-88131 Lindau, Bavaria\\[1mm]
        vasili.baranov@gmail.com
        }
    }
\fi

\usepackage{url}
\makeatletter
\def\Uslash{\mathbin{\mathchar`\/}\@ifnextchar{/}{\kern-.15em}{}}
\g@addto@macro\UrlSpecials{\do \/ {\Uslash}}
\def\Ucolon{\mathbin{\mathchar`:}\@ifnextchar{/}{\kern-.1em}{}}
\g@addto@macro\UrlSpecials{\do : {\Ucolon}}
\makeatother

\begin{document}

\twocolumn[{\csname @twocolumnfalse\endcsname

\maketitle  

\begin{abstract}
\noindent
In this paper, we focus on improving binary 2D instance segmentation to assist humans in labeling ground truth datasets with polygons. Humans labeler just have to draw boxes around objects, and polygons are generated automatically. To be useful, our system has to run on CPUs in real-time. The most usual approach for binary instance segmentation involves encoder-decoder networks. This report evaluates state-of-the-art encoder-decoder networks and proposes a method for improving instance segmentation quality using these networks. Alongside network architecture improvements, our proposed method relies upon providing extra information to the network input, so-called ``extreme points'', i.e. the outermost points on the object silhouette. The user can label them instead of a bounding box almost as quickly. The bounding box can be deduced from the extreme points as well. This method produces better IoU compared to other state-of-the-art encoder-decoder networks and also runs fast enough when it is deployed on a CPU.

\end{abstract}

\subsection*{Keywords}
Extreme points, IoU, Encoder-Decoder, Instance binary segmentation.

\vspace*{1.0\baselineskip}
}]


\section{Introduction}
Visual recognition tasks are currently active research topics in the areas of autonomous driving, biomedical image processing, and scene understanding. To accomplish automatic visual recognition based on deep learning, training a deep neural network to learn to extract features from images is essential. To do this there is a need for a lot of annotated training data. For this, manual labeling is used, which is very time-consuming. It includes locating manually the precise positions, drawing bounding boxes, assigning labels, and drawing polygons around the objects. Hence, this process is tardy, requires a lot of manpower for labeling enormous data and human labelers are prone to errors as well. The objective of this paper was to develop a fast and precise system that can perform binary instance segmentation so that human labelers are significantly assisted in the task of manual 2D semantic segmentation of road-scene objects.

\begin{figure}[tb]
	\centering
	\includegraphics[width=7cm,height=10cm, keepaspectratio]{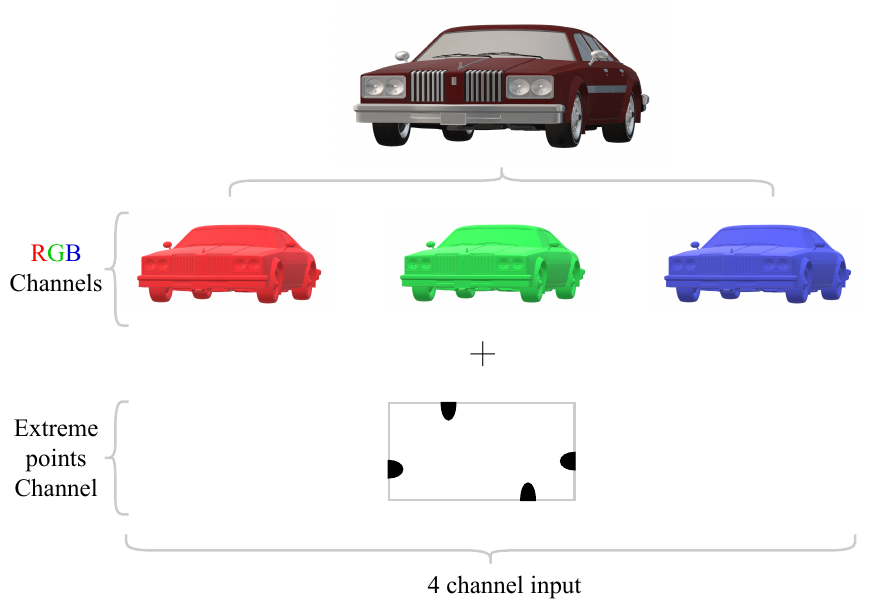}
	\caption{Extreme points channel along with RGB channels.}
	\label{fig:Extreme_points}
\end{figure}

Instance segmentation task is simultaneously solving both tasks of object detection and semantic segmentation \cite{khe17, hua19}, it is needed to locate in an image every object from a predefined set of classes as well as to give a binary mask for each of these objects. Other flavor of instance segmentation is binary segmentation \cite{ron15}, here it is only a single object that has to be segmented, i.e. pixels are classified as foreground and background because the segmentation happens only in the object bounding box, which has to be obtained from a detection algorithm or manual annotation.


Initial manual annotation of data is crucial for training deep learning models. A typical manual annotation is drawing polygons around each object of interest. It will be possible to significantly assist human labelers in the task of manual 2D semantic segmentation (i.e. specifying polygons around each object of interest in a scene), if we develop a fast and accurate system-a neural network-for binary instance segmentation (foreground/background). The network shall produce a silhouette of an object (e.g., a car or a person) when a user specifies a corresponding 2D bounding box.

In order to speedup manual drawing of polygons around objects, this paper investigates state-of-the-art encoder-decoder architectures for binary instance segmentation and proposes a method with extreme points as shown in \figu \ref{fig:Extreme_points}.  By providing only the bounding box or extreme points (see below), one can generate polygons through our binary instance segmentation algorithm inside the
\ifanonymous
     \textbf{labeling tool developed by a well known company}
\else
    C.LABEL labeling tool developed by CMORE Automotive GmbH 
\fi
(where the present research was performed). The proposed method significantly assists human labelers in the task of manual 2D semantic segmentation of road-scene objects by automatically generating polygons around each object of interest in an image. This paper also focuses on deploying the proposed method on a normal CPU rather than using it in a GPU (human annotators often do not have access to GPUs) such that it shall run fast enough and also occupy little hard drive and RAM space. In this way, it can be easily deployed with the labeling tool. Our target was inference time $\leq$ 200 ms on a CPU.\\


\section{Network Architecture}
Our approach is built upon the encoder-decoder architecture with skip connections, U-Net\cite{ron15}. The U-Net architecture consists of a encoder path to capture the context, a symmetrical decoder path that enables accurate object detection, and skip connections \cite{ron15}. The main idea of encoder-decoder networks is to supplement a usual encoder network by successive layers, where pooling operators are replaced by upsampling operators \cite{ron15}. This results in increase in the resolution than the encoded representation. The high-resolution features from the encoder path are additionally combined with upsampled decoder features \cite{ron15}. The decoder network can learn to produce more accurate output based on the information from skip connections \cite{ron15}. The output of the model is the binary mask of the object in an image.

\begin{figure}[tb]
	\centering
	\includegraphics[width=7cm,height=10cm, keepaspectratio]{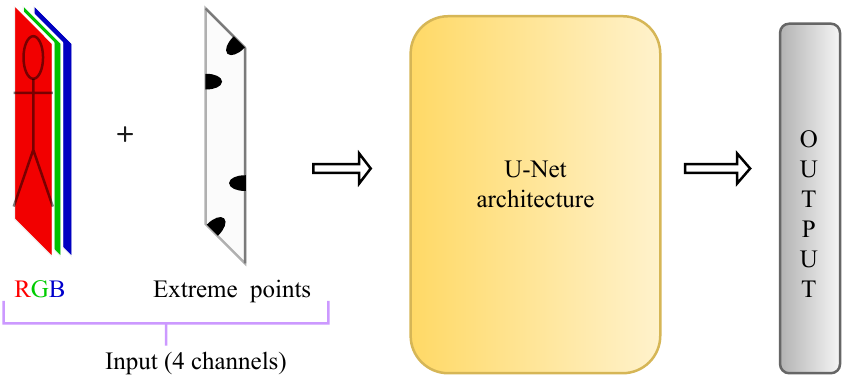}
	\caption{Extreme points input in the U-Net architecture.}
	\label{fig:proposed_network}
\end{figure}

We modified the U-Net architecture in several ways: (i) by using depth-wise separable convolutions instead of convolutions \cite{cho17}; (ii) by using residual blocks \cite{hez16} instead of normal convolutions; (iii) by using dense blocks \cite{hua17} instead of normal convolutions; (iv) by using contextual convolutions \cite{ono18}; and (v) by using semantically similar skip connections \cite{zho18} in order to improve the performance with U-Net. 
These modifications didn't lead to significant segmentation quality improvements under our inference time restriction ($\leq$ 200 ms on a CPU), that's why we explored providing more input data to the network.

To provide more input data to the network, we pass extreme points information \cite{man18} to the network along with the normal RGB input (RGB is a standard choice) (cf. \figu \ref{fig:Extreme_points} and \figu \ref{fig:proposed_network}). The main intention is to increase the precision of the segmented masks by letting the network use extra information. The extreme points are left, right, top, bottom --most pixels of the object, i.e. the points where the object touches its bounding box. The technique of using extreme points for binary segmentation was introduced in \cite{man18}. That paper demonstrates the networks accepting RGB channels + extreme points perform better than equivalent networks that accept RGB channels only. Extreme points are represented with a binary mask in the additional channel. In the original paper, a 2D Gaussian is applied to each of these points in the binary mask. In this work we used drawing circle around every point. Marking extreme points is as fast or faster as marking bounding boxes. That is why we can rely on this information in our work.

The use of an additional channel along with the input 3-channel image in the U-Net architecture \cite{ron15} is shown in \figu \ref{fig:proposed_network}. The convolution architecture is the same as U-Net but the basic filter depth used is $f = 16$ in the first level. The filter size is progressively increased in the encoder as: $f$, $2f$, $4f$, $8f$, and $16f$. Similarly, the filter size in decoder is decreased from $16f$ as $8f$, $4f$, $2f$, and $f$. The input size is proportionally decreased in every level in the encoder using max pooling operation as $128\times128$, $64\times64$, $32\times32$, $16\times16$, and $8\times8$. In the same way in decoder the input size is increased using upsampling operation from $8\times8$ as $16\times16$, $32\times32$, $64\times64$, and $128\times128$. The final output size is $128\times128$.

\section{Training}

The network is trained from scratch with a total 83,403 instances for training and 15,926 instances for validation from the Cityscapes dataset \cite{cor16}, simultaneously for the 9 most prevalent classes from the dataset (``car'', ``traffic sign'', ``bicycle'', ``person'', ``rider'', ``motorcycle'', ``traffic light'', ``truck'', and ``bus''). The instances are re-scaled to $128 \times 128$ pixels before feeding into the network.

We choose Intersection Over Union (IoU) and ``border error'' as the evaluation metric. The ``border error'' is the per-pixel distance from the predicted object boundary to the ground truth boundary. For comparison, we calculate different flavors of IoU, such as average IoU (aIoU- the IoU that is calculated for each validation batch and then averaged for all the batches to produce aIoU of that epoch), mean IoU (mIoU- the IoU that is calculated by averaging class-based IoUs), and instance IoU (iIoU- the IoU that is calculated by averaging IoUs calculated per single instance) as introduced in \cite{cir19}. The extreme points channel is created during training as an additional channel with a binary mask and passed to the network along with a RGB channels. We apply the following data augmentations to instances during training: flip and rotation. We use several loss functions: differentiable IoU loss as introduces in \cite{cir19}, combination of the IoU loss and binary cross entropy \cite{igl18}, and our custom loss function that approximates the average distance error (see below). Our custom loss function showed better results than other loss functions. The average distance error is the line integral $\int_{border} EDT(s) ds / L$, where EDT is the Euclidean Distance Transform and L is the ground truth boundary length. We approximate L as $\sqrt{A_g}$, and the approximate average distance error $\overline{\Delta d}$ is given by    
\begin{equation}
\overline{\Delta d} =\ \frac{ A_u - A_i}{\sqrt{A_g}},
\label{average-metric-loss}
\end{equation}
\begin{equation} 
\overline{\Delta d} =\ \frac{\abs{y \; \cup \; y ^\prime}\; -\; \abs{y \; \cap \; y ^\prime}}{\sqrt{A_g}},
\label{average-metric-loss1}
\end{equation}
where $A_u$ and $A_i$ is the area of union and area of intersection of the groundtruth and predicted masks respectively,  $y$ is the predicted pixel probabilities, $y ^\prime$ is the groundtruth binary labels, $A_g$ is the area of the groundtruth boundary, $\cap$ is the intersection operation and $\cup$ is the union operation.


During validation and during training, the average IoU (aIoU) is calculated as the metric. After a network is trained and the best epoch is selected, we calculated the mean IoU (mIoU), instance IoU (iIoU), and per-pixel distance. The aIoU, mIoU, iIou, and per-pixel distance (border error) are calculated in images scaled to the network input size ($128\times128$ pixels). Data augmentation techniques are used only during training (metrics are calculated over the validation dataset after each epoch). During evaluation (testing the model after training) over the validation dataset, data augmentation techniques are switched off to calculate per-pixel distance, mIoU, and iIoU. 
The discussed approaches are trained on one Nvidia GeForce GTX 1080 Ti GPU and are deployed during inference on an Intel(R) Core(TM) i5-6300U CPU for measuring the time taken for segmenting objects.

\section{Results}

The network is trained from scratch with the ``Adam'' optimizer and the learning rate of $1\times10^{-3}$ using the average distance loss function. In this method, the network is trained for 100 epochs with a batch size of 32, while we observed in experiments that training starts to converge after 18--20 epochs and the best results are obtained within this range. We have used early stop criteria and selected the best model. The per-pixel distance is calculated over the validation dataset after training, using the best model from the network.

\begin{figure}[tb]
	\centerline{\includegraphics[width=8.0cm,height=10cm, keepaspectratio]{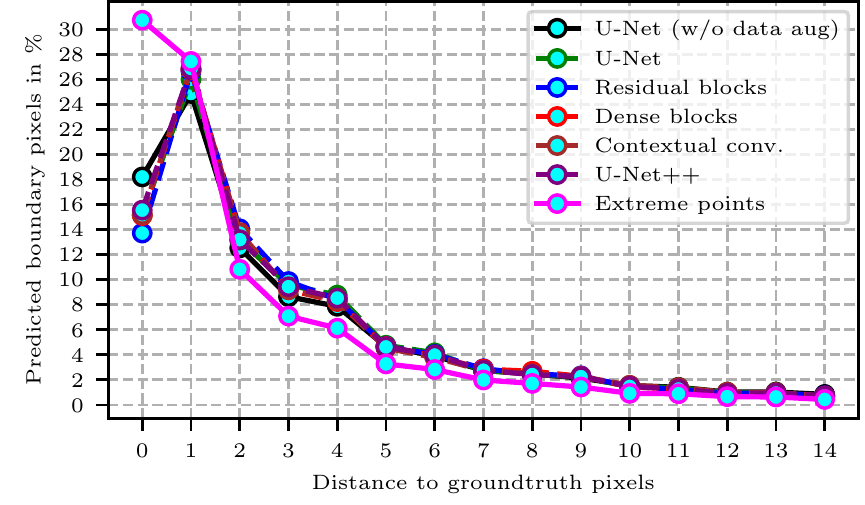}}
	\caption{Comparison of per-pixel distance distributions for the discussed approaches.}
	\label{fig:comparison_discussed_approaches_data_aug_ppd}
\end{figure}

The comparison of the border error in different U-Net variants studied is shown in \figu \ref{fig:comparison_discussed_approaches_data_aug_ppd}. The approach with extreme points has a higher probability of error at 0 px, which implies a higher accuracy of the model.
In Table \ref{tab1}, the comparison of aIoU for the best models over the validation dataset shows that the approach with extreme points performs better than other approaches used in this work with, \textbf{89.16\%} IoU. 
The mIoU and iIoU, which are calculated after training (data augmentation is switched off) are \textbf{90.65\%} and \textbf{87.25\%} for the network with extreme points, respectively. These results evidently show that adding extreme points leads to higher quality of segmentation. The comparison of per-class mIoUs of the most interesting classes for the discussed U-Net variants trained with data augmentations and the average distance loss is shown in Table \ref{tab2}. The table also demonstrates that the approach with extreme points produces the best results for all the displayed classes.
In \figu \ref{fig:results} one can see a qualitative evaluation with examples of the important car and person classes.

\setlength{\tabcolsep}{1.000pt}
\renewcommand{\arraystretch}{1.0}
\begin{table}[htbp]
\small
	\begin{center}
		\newcolumntype{P}[1]{>{\centering\arraybackslash}p{#1}}
		\newcolumntype{M}[1]{>{\centering\arraybackslash}m{#1}}
		\begin{tabular}{|M{1.8cm}|M{1.3cm}|M{1.3cm}|M{1.3cm}|M{1.5cm}|}
			\hline
			
			\textbf{Approach} & \textbf{\textit{aIoU}}(\si{\%})& \textbf{\textit{iIoU}}(\si{\%})& \textbf{\textit{mIou}}(\si{\%})&  \textbf{Inference Time}
			(\si{ms})\\
			\hline
			\textbf{U-Net} & 84.45 & 86.38 & 83.36&  \textasciitilde{}140\\
			\hline
			\textbf{Residual blocks} & 85.11 & 86.47 & 83.88&  \textasciitilde{}180\\
			\hline
			\textbf{Dense blocks} & 85.88 & 86.99 & 84.92&  \textasciitilde{}320\\
			\hline
			\textbf{Contextual conv.} & 85.12 & 86.75 & 83.70&  \textasciitilde{}240\\
			\hline
			\textbf{U-Net++} & 85.47 & 86.10 & 82.73&  \textasciitilde{}380\\
			\hline
			\textbf{Extreme points} & \textbf{89.16} & \textbf{90.65} & \textbf{87.25}&  \textasciitilde{}140\\
			\hline
		\end{tabular}
	\end{center}
	\caption{Comparison of performance metrics of U-Net variants trained with the average distance loss}
	\label{tab1}
\end{table}

\begin{table}[htbp]
\small
	\begin{center}
		\newcolumntype{P}[1]{>{\centering\arraybackslash}p{#1}}
		\newcolumntype{M}[1]{>{\centering\arraybackslash}m{#1}}
		\begin{tabular}{|M{1.7cm}|M{0.8cm}|M{0.8cm}|M{0.95cm}|M{0.9cm}|M{0.8cm}|M{1cm}|}
			\hline
			
			\textbf{Approach} & \textbf{\textit{car}}& \textbf{\textit{bus}}& \textbf{\textit{bicycle}} & \textbf{\textit{person}} & \textbf{\textit{rider}} & \textbf{\textit{motor-
					cycle}}\\
			\hline
			\textbf{U-Net} & 90.39 & 90.22 & 77.09 & 80.43 & 75.30 & 76.98 \\
			\hline
			\textbf{Residual blocks} & 90.42 & 89.91 & 76.68 & 80.97 & 75.58 & 76.63  \\
			\hline
			\textbf{Dense blocks} & 90.52 & 91.01 & 77.20 & 81.13 & 75.77 & 77.04  \\
			\hline
			\textbf{Contextual conv.} & 90.57 & 91.07 & 76.72 & 80.41 & 75.95 & 75.50  \\
			\hline
			\textbf{U-Net++} & 90.85 & 91.25 & 77.27 & 81.29 & 75.66 & 76.69  \\
			\hline
			\textbf{Extreme points} & \textbf{92.49} & \textbf{93.26} & \textbf{81.16} & \textbf{84.92} & \textbf{80.08} & \textbf{81.54}  \\
			\hline
		\end{tabular}
	\end{center}
	\caption{Comparison of per-class mIoUs for U-Net variants trained with the average distance loss}
	\label{tab2}
\end{table}
We found out that all the architecture modifications except the extreme points one perform almost identically to the basic U-Net architecture. If inference time is kept fixed, changes in the architecture do not seem to be able to significantly improve network accuracy. But, providing additional user input (extreme points) allowed the network to produce significantly better results, on the other hand.

\setlength{\mylenth}{2.2cm}
\begin{figure}[tbp]
	\centering
	\subfigure{
		\includegraphics[height=\mylenth]{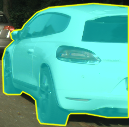}
	}
	\subfigure{
		\includegraphics[height=\mylenth]{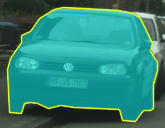}
	}
	\subfigure{
		\includegraphics[height=\mylenth]{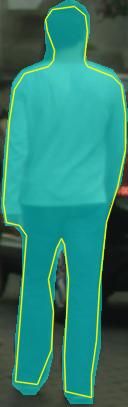}
	}
	\subfigure{
		\includegraphics[height=\mylenth]{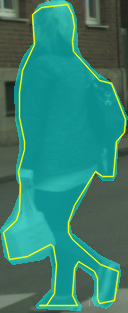}
	}
	\caption{Results of our method for the car and person class (turquoise surface is ground truth,  yellow contour the prediction).}
	\label{fig:results}
\end{figure}

\section{Conclusion}

Our method with extreme points in the U-Net \cite{ron15} architecture achieves good performance and clearly outperforms the other approaches discussed in this paper. Our approach achieves 3.7 points gain of aIoU, 4.5 points gain of iIoU and 4.5 points gain of mIoU over the basic U-Net architecture. The inference time for binary instance segmentation is the same for the basic U-Net and U-Net with extreme points approaches, which is about 140 ms. In this work, we also introduced a new custom loss function that matches the per-pixel error slightly better than the differentiable IoU loss. This improved architecture is already being integrated into the manual annotation tool of
\ifanonymous
    \textbf{well known company} with deep learning assistance capabilities.
\else
    CMORE Automotive GmbH with deep learning assistance capabilities, C.LABEL.
\fi
 For commercial use, the network was retrained on an internal dataset.

One possible approach to further improve the accuracy of segmentation results is to use Generative Adversarial Networks (GANs) \cite{goo14}, where the generator shall produce a binary segmentation and the discriminator shall distinguish true segmentation masks from generated ones. The discriminator shall essentially learn the evaluation metric on its own.


\end{document}